\documentclass{article} % For LaTeX2e
\usepackage{iclr2023_conference,times}

\usepackage{amsmath,amsfonts,bm}

\def\eqref#1{equation~\ref{#1}}
\def\1{\bm{1}}

\DeclareMathAlphabet{\mathsfit}{\encodingdefault}{\sfdefault}{m}{sl}
\SetMathAlphabet{\mathsfit}{bold}{\encodingdefault}{\sfdefault}{bx}{n}

\DeclareMathOperator{\sign}{sign}

\usepackage{booktabs}
\usepackage{multirow, makecell}
\usepackage{subfigure}
\usepackage{caption}
\usepackage{textcomp}
\usepackage{color, colortbl}
\usepackage{mathtools}
\usepackage{appendix}
\usepackage{bm}
\usepackage{bbm}
\usepackage{wrapfig}
\usepackage{xspace}
\usepackage{enumitem}
\usepackage{lineno}

\usepackage{hyperref}
\hypersetup{
colorlinks=true,
linkcolor=red,
citecolor=green
}
\usepackage{url}

\definecolor{gray}{RGB}{120, 120, 120}
\definecolor{darkred}{RGB}{135, 12, 12}
\definecolor{lightblue}{RGB}{0, 128, 192}
\definecolor{blue2}{RGB}{0, 64, 120}
\definecolor{reviewer2}{RGB}{0, 0, 255}
\definecolor{reviewer3}{RGB}{255, 0, 255}
\definecolor{reviewer4}{RGB}{240, 134, 80}

\newcommand{\Ra}[1]{{#1}}
\newcommand{\Rb}[1]{{#1}}
\newcommand{\Rc}[1]{{#1}}
\newcommand{\Rd}[1]{{#1}}

\title{%\centering
Compositional Prompt Tuning with Motion \\ Cues for Open-vocabulary Video Relation \\ Detection
}

\author{Kaifeng Gao$^{1}$, Long Chen$^{2}$\thanks{Long Chen is the corresponding author. Part of the work was done when Kaifeng Gao served as a visiting Ph.D. student at Singapore Management University.}, ~Hanwang Zhang$^3$, Jun Xiao$^1$, Qianru Sun$^4$ \\
$^1$Zhejiang University, $^2$The Hong Kong University of Science and Technology \\
$^3$Nanyang Technological University, $^4$Singapore Management University \\
\texttt{$^1$\{kite\_phone,junx\}@zju.edu.cn, $^2$zjuchenlong@gmail.com} \\
\texttt{$^3$hanwangzhang@ntu.edu.sg, $^4$qianrusun@smu.edu.sg}
}

\newcommand{\eg}{\emph{e.g.}}
\newcommand{\ie}{\emph{i.e.}}
\iclrfinalcopy % Uncomment for camera-ready version, but NOT for submission. 
\begin{document}

% \linenumbers

\maketitle

%% 一定要举例什么是visual relation detection，因为ICLR的审稿人很可能没做过这个task。
% 1. Video relation detection is a challenging task: introduction. 省掉vqa之类的mention，直接放到引用里。challenge在哪里?标注！（manually annotation)

% 2. 所以需要open vocabulary：定义一下。这里才是真正要highlight的challenge：1）如何利用有限的label去transfer knowledge到关联的label。比如有某某方法。举例子(以前的排列组合 zero-shot)。2）但还有很多是根本没见过的，这个就无解。只能用到大模型的encyclopedic knowledge. 举例子。

% 3. prompt介绍 参考一下 贝尔师兄 的论文。

% 4. 但prompt for video的问题. 1) composition: subject and object. 而不是你说的diverse。2）temporal diversity 跟你说的第二点一样。
% aims to detect novel object categories beyond the training set

\begin{abstract}
Prompt tuning with large-scale pretrained vision-language models empowers open-vocabulary predictions trained on limited base categories, \eg, object classification and detection. In this paper, we propose \emph{compositional} prompt tuning with \emph{motion cues}: an extended prompt tuning paradigm for compositional predictions of video data. In particular, we present Relation Prompt (RePro) for Open-vocabulary Video Visual Relation Detection (Open-VidVRD), where conventional prompt tuning is easily biased to certain subject-object combinations and motion patterns. To this end, RePro addresses the two technical challenges of Open-VidVRD: 1) the prompt tokens should respect the two different semantic roles of subject and object, and 2) the tuning should account for the diverse spatio-temporal motion patterns of the subject-object compositions. Without bells and whistles, our RePro achieves a new state-of-the-art performance on two VidVRD benchmarks of not only the base training object and predicate categories, but also the unseen ones. Extensive ablations also demonstrate the effectiveness of the proposed compositional and multi-mode design of prompts. Code is available at \url{https://github.com/Dawn-LX/OpenVoc-VidVRD}.
\end{abstract}

\section{Introduction}
\begin{wrapfigure}{r}{45ex}
    \vspace{-3ex}
    \centering
    \includegraphics[width=\linewidth]{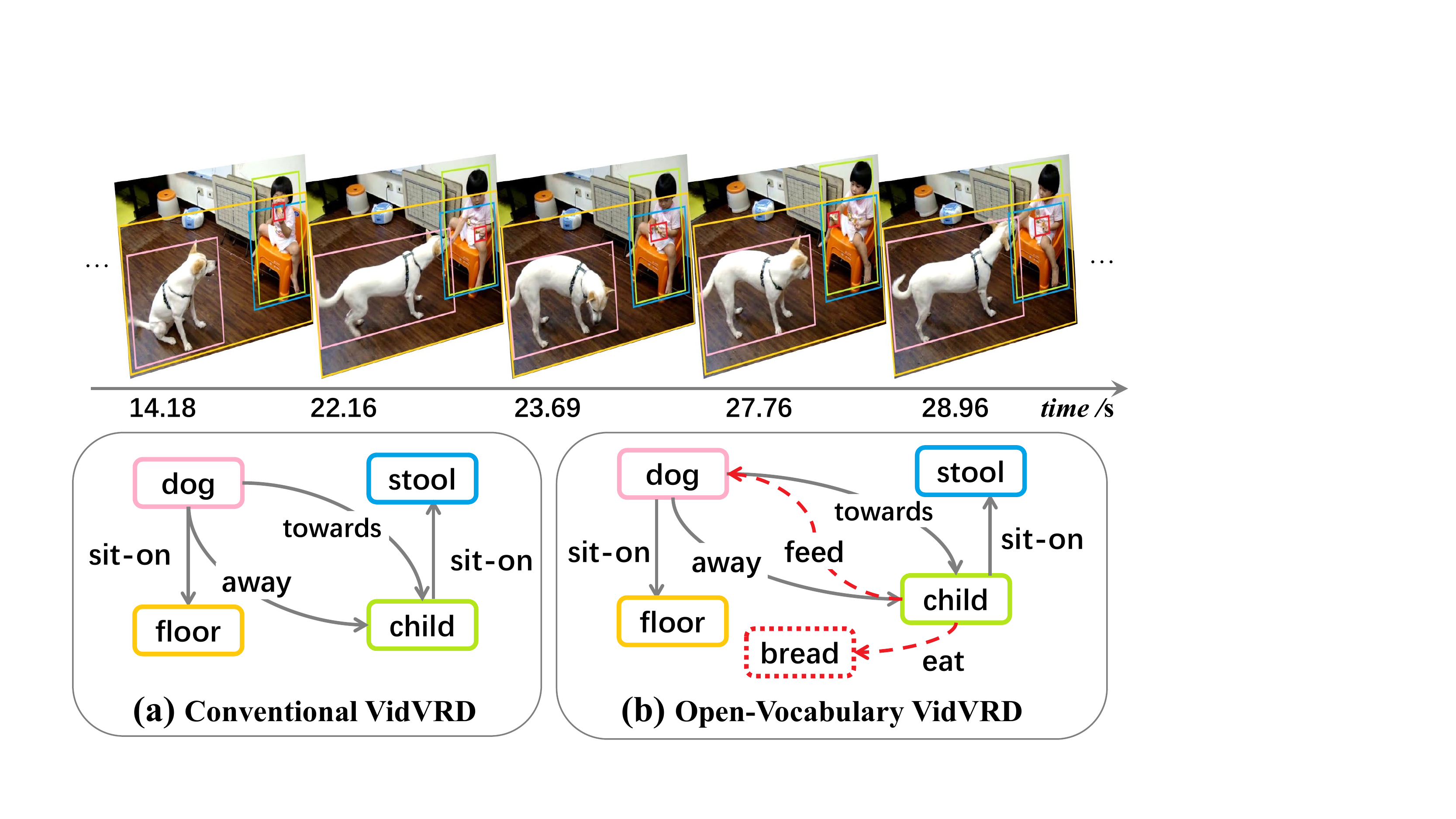}
    \vspace{-2ex}
    \caption{Examples of VidVRD. The relation graphs are w.r.t the whole video clip. Dashed lines denote unseen \emph{new} categories in the training data.}\label{fig_fig1}
    \vspace{-2ex}
\end{wrapfigure}

%% 一定要举例什么是visual relation detection，因为ICLR的审稿人很可能没做过这个task。
% 1. Video relation detection is a challenging task: introduction. 省掉vqa之类的mention，直接放到引用里。challenge在哪里?标注！（manually annotation)

% 2. 所以需要open vocabulary：定义一下。这里才是真正要highlight的challenge：1）如何利用有限的label去transfer knowledge到关联的label。比如有某某方法。举例子(以前的排列组合 zero-shot)。2）但还有很多是根本没见过的，这个就无解。只能用到大模型的encyclopedic knowledge. 举例子。

% 3. prompt介绍 参考一下 贝尔师兄 的论文。

% 4. 但prompt for video的问题. 1) composition: subject and object. 而不是你说的diverse。2）temporal diversity 跟你说的第二点一样。

% Video visual relation detection (VidVRD) is a challenging task: it aims to detect the pair-wise visual relationships between object tracklets in videos, \eg, \texttt{dog}-\texttt{towards}-\texttt{child} as shown in Figure~\ref{fig_fig1}.

%##################### 1 ######################
Video visual relation detection (VidVRD) aims to detect the visual relationships between object tracklets in videos as ⟨subject, predicate, object⟩ triplets~\citep{shang2017video,Chen2021Social,chen2023video,gao2021video,gao2022classification}, \eg, \texttt{dog}-\texttt{towards}-\texttt{child} shown in Figure~\ref{fig_fig1}. 
% 这种can be decomposed to many的写法感觉不太好。 感觉还是强调一下，相比于image，由于不同时间长度的选取，video relation的类别会有不同level/scale的变化，之类的.gkf:不要强调 decompose 这一点,容易让人confuse
Compared to its counterpart in still images~\citep{chen2019counterfactual,li2022devil,li2022nicest,li2022integrating,li2022rethinking}, due to the extra temporal axis, there are usually multiple relationships with different temporal scales, and a subject-object pair can have several predicates with ambiguous boundaries. For example, as shown in Figure~\ref{fig_fig1}, the action \texttt{feed} of \texttt{child} to \texttt{dog} co-occurs with several other predicates (\eg, \texttt{away}, \texttt{towards}). This characteristic makes VidVRD have more plentiful and diverse  relations between objects than its image counterpart. As a result, it is impractical to collect sufficient annotations for all categories for VidVRD. Therefore, to make VidVRD practical, we should know how to generalize the model, trained on limited annotations, to \emph{new} object and predicate classes \emph{unseen} in training data. 

To this end, we propose a new task: \emph{Open-vocabulary} VidVRD (Open-VidVRD). In particular, ``open'' does not only mean unseen relationship combinations, \eg, \texttt{dog-sit\_on-floor}, but also unseen objects and predicates, \eg, \texttt{bread} and \texttt{feed}, as shown in Figure~\ref{fig_fig1}. Recent works on such generalization only focus on the unseen combinations~\citep{Chen2021Social,shang2021video} in VidVRD, or zero-shot transfer among semantically related objects in zero-shot object detection~\citep{Huang2022Robust}, \eg, the seen \texttt{dog} class can help to recognize the unseen \texttt{wolf}. However, they fail to generalize to the categories totally unrelated to the  limited seen ones, where the transfer gap is unbridgeable, \eg, \texttt{bread} in testing has no visual similarity with \texttt{dog} and \texttt{child} in training. 
% \redtxt{We show in the experiments Sec.~\ref{sec_exps} that our method has the zero-shot transferability to object and predicate categories totally unseen in the training set.}

% \redtxt{[We show in the experiments xxxxx.]}

% For visual relations totally unrelated to the seen limited ones, the transfer gap is unbridgeable. For example, detecting \texttt{frog-kiss-frog} is impossible when the training data only covers indoor activities.
% % xxxx xxx .% 一定要找到unseen根本不存在seen的visual feature的例子 gap足够大的例子。

% unseen in the training
% set. To achieve the real ability of open-vocabulary recognition, we choose to utilize the encyclopedic
% knowledge from large-scale pre-trained visual-language models (VLM).

% fig1中明显画出train&test

\begin{figure}[t]
    \vspace{-2ex}
    \centering
    \includegraphics[width=\linewidth]{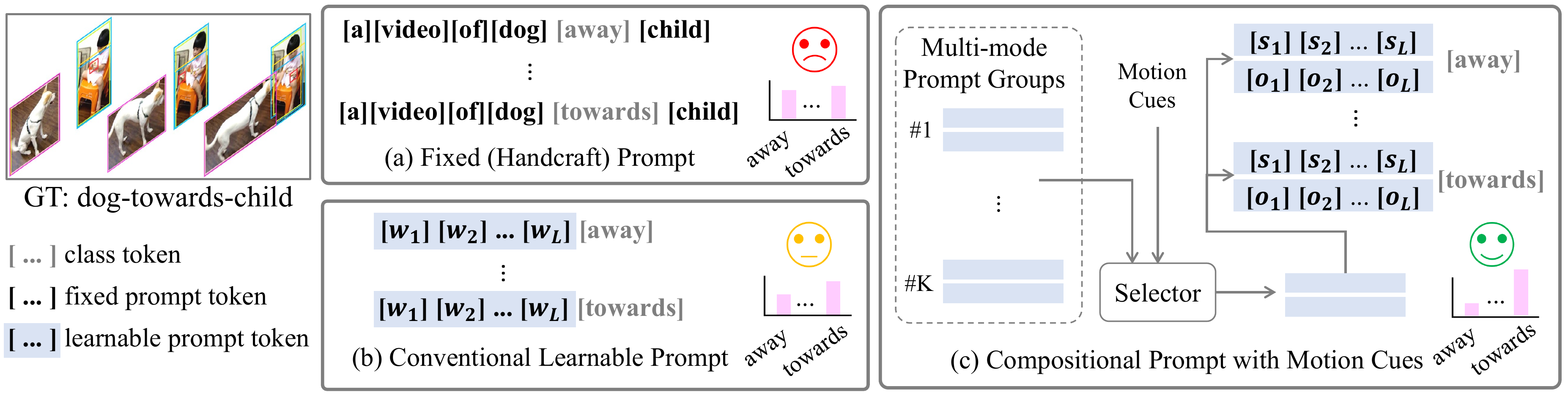}
    \vspace{-3ex}
    \caption{Comparisons of different prompt tuning methods for Open-VidVRD.}
\label{fig_prompt}
\vspace{-3ex}
\end{figure}

Thanks to the encyclopedic knowledge acquired by large vision-language models (VLMs) pre-trained on big data~\citep{radford2021learning,li2022align}, we can achieve open-vocabulary relation detection with only training data of limited base categories. To bridge the gap between the pre-trained and downstream tasks without extra fine-tuning the whole VLM model, a trending technique named \textbf{prompt tuning} is widely adopted~\citep{liu2021pre,jin2022good,zhou2022learning}. For example, we can achieve zero-shot relation classification for the tracklets pair in Figure~\ref{fig_prompt}. We first crop the object tracklet regions in the video, and feed them into the visual encoder of VLM to obtain corresponding visual embeddings. Then we use a simple prompt like ``a video of [CLASS]'', feed it to the VLM's text encoder to obtain the text embedding, and classify the object based on the similarities between visual and text embeddings. Based on the tracklet classification results, for the example of the pair \texttt{dog} and \texttt{child}, we can craft a prompt like ``a video of dog [CLASS] child'', as shown in Figure~\ref{fig_prompt}(a), and similarly classify their predicates based on the predicate text embeddings. Furthermore, we can replace the fixed prompt tokens with learnable continuous tokens, as shown in Figure~\ref{fig_prompt}(b), known as prompt representation learning, which has been widely applied to open-vocabulary object detection~\citep{gu2021open,du2022learning,ma2022open}.
% which has been widely applied to many vision tasks, \eg, image classification~\citep{zhou2022learning,zhou2022conditional} and open-vocabulary object detection~\citep{gu2021open,du2022learning,ma2022open}. 

% for the pair \texttt{dog} and \texttt{child} shown in Figure~\ref{fig_fig1} based on a simple prompt like :
% %%% 【NOTE】 <-- 我们的prompt里面没有 dog 和 child的信息, 我们在后面要提到这一点，因为要做Open-Vocabulary， 所以不能在prompt里面 conditioned on object category, 这样会over-fit 到 base category
% we first crop the tracklet regions of child and dog in the video, and feed them into the visual encoder of VLM to obtain corresponding visual embeddings. Then we input the prompt into the text encoder and extract text embeddings for all predicate classes. Finally, we classify the predicates based on the similarities between the corresponding visual and text embeddings. 

%%% TODO this paragraph is not match to Figure.1

Learning the prompt representation is actually introducing some priors for describing the context of target classes, and it excludes some impossible classes with the constraint of the context. 
However, the prompt representations (either handcrafted or learned) 
% (either handcrafted as Figure~\ref{fig_prompt}(a) or learned as Figure~\ref{fig_prompt}(b))
in above approaches are monotonous and static, and learning the prompt sometimes might break the ``open'' knowledge due to overfitting to the base category training data. %%【还是决定不加 break the ``open'' knowledge这一点?，因为我下面的两个\item 里面没有说到overfit 这一点。
Modeling the prompt representation for video visual relations has some specific characteristics that need to be considered:
\begin{itemize}[leftmargin=15pt]
    \vspace{-1ex}
    \item \textbf{Compositional}: The prompt context for predicates is highly related to the semantic roles of subject and object. A holistic prompt representation might be sub-optimal for predicates. For example, as shown in Figure~\ref{fig_fig1}, even the same predicate (\texttt{sit\_on}) in different relation triplets (\texttt{dog-sit\_on-floor} and \texttt{child-sit\_on-stool}) have totally different visual context.
    \item \textbf{Motion-related}: Predicates with different motion patterns naturally should be prompted with different context tokens. The naive prompt representation fails to consider the spatio-temporal motion cues of tracklet pairs. For example, the predicate \texttt{towards} shown in Figure~\ref{fig_fig1} can be prompted as ``a relation of [CLASS], moving closer''. In contrast, \texttt{eat} and \texttt{sit\_on} can be prompted as ``a relation of [CLASS], relative static''.
     \vspace{-1ex}
\end{itemize}

In this paper, we propose a compositional and  motion-based \textbf{Re}lation \textbf{Pro}mpt learning framework: RePro, as shown in Figure~\ref{fig_prompt}(c). \Ra{To deal with the \textbf{compositional} characteristic of visual relations, we set compositional prompt representations specified with subject and object respectively}. \Ra{With this design, we can model the prompt context w.r.t. semantic roles (\ie, \texttt{subject} or \texttt{object})}. For example, a possible prompt can be ``sth. doing [CLASS]''  for the subject and ``sth. being [CLASS]'' for the object. 
%%%%%%%###############################
% cmpositional prompt 得到的 concatenated text embedding 要求我们的visual embedding 也是subject-object concated 的， 这就相当于在做relation 分类的时候，考虑了 subject 处于这个relation的概率 乘以 object 处于这个relation的概率（还有一个 sqrt(2)的缩放）。
%%%%%%%###############################
% To consider the \textbf{motion-related} context of predicates, we design multi-mode prompt groups, as shown in Figure~\ref{fig_prompt}(c), where multi-mode means \emph{with multiple motion patterns}. Each group has the compositional prompts for subject and object, and we select a proper group according to the motion cues (patterns) in the subject-object tracklet pairs (cf. Sec.~\ref{Sec_RelCls}). 
\Ra{To consider the \textbf{motion-related} characteristic of predicate contexts, we design multi-mode prompt groups, where each group (\ie, each mode) is assigned with a certain motion pattern, and has its own compositional prompts for the subject and object.} During the implementation, we select a proper group according to the motion cues (patterns) in the subject-object tracklet pairs (cf. Sec.~\ref{Sec_RelCls}). 
Compared to some prompt tuning works which focus on category-based context~\citep{zhou2022learning} or \Ra{instance-conditioned context~\citep{zhou2022conditional,ni2022expanding}}, our motion-cue-based grouping has better cross-category generalization ability, and can avoid the over-fitting to base categories. We evaluate our RePro on the VidVRD~\citep{shang2017video} and VidOR~\citep{shang2019annotating} benchmarks. Our experiment results show that RePro trained with only the samples of \emph{base} relation categories has a good generalizability to detect \emph{novel} relations, and achieves the new state-of-the-art. For example, it outperforms the top-performing method, \ie, VidVRD-II~\citep{shang2021video}, by $2.54\%$ and $3.91\%$ absolute mAP for SGDet and SGCls settings, respectively.

% Extensive ablations also have demonstrated the effectiveness of our compositional and motion-based prompt design.

Our contributions in this paper are thus three-fold. 1) A new open-vocabulary setting for video visual relation detection task, \ie, Open-VidVRD. 2) A compositional prompt representation learning method that models the prompt contexts for the subject and object separately. 3) A motion-cue-based multi-mode prompt groups that achieve a strong generalization ability.
% and avoid the over-fitting to base categories.

% $\bullet$ We propose a new open-vocabulary setting for video visual relation detection task, \ie, Open-VidVRD.

% $\bullet$ We propose a compositional prompt representation learning strategy, which better model the prompt context for subject and object respectively.

% $\bullet$ We propose the motion pattern based prompt selection strategy, which achieve better cross-category generalize ability and avoid the over-fitting to base categories.

\section{Related Work}

\textbf{Video Visual Relation Detection (VidVRD)} was defined in~\cite{shang2017video,shang2019annotating} together with the proposals of the VidVRD and VidOR benchmarks. 
The task aims to spatio-temporally localize visual relations between object tracklets.
Existing methods mainly focus on modeling better visual or spatio-temporal contexts~\citep{qian2019video,shang2021video,cong2021spatial}, and detecting visual relations with more granularity either by sliding windows~\citep{liu2020beyond} or temporal grounding~\citep{gao2022classification}. \Ra{They mainly worked on the pre-defined (closed) sets of object and predicate categories. In contrast, our work is the first one to study the open-vocabulary VidVRD setting, where some object and predicate categories are unseen in the training set.}

\Ra{
\textbf{Zero-Shot Setting in Image and Video VRD}. Existing VRD works, either in image domain~\citep{tang2020unbiased,kan2021zero} or video domain~\citep{shang2021video}, only achieve zero-shot transfer on the unseen triplet combinations, where the objects and predicates are seen in the training set. They ignore the model's generalization ability to unseen object/predicate categories. There is one concurrent work~\citep{he2022towards} proposes the open-vocabulary setting in image VRD. However, they put the main emphasis on unseen object categories. Different from them, RePro generalizes the model to recognize both object and predicate categories totally unrelated to the seen training ones.
}

% which had been discussed in Section 2.  
% They mainly worked on the pre-defined visual relations with closed sets of object and predicate categories (in both training and testing phases), and only achieve  zero-shot transfer on the unseen combinations, while ignore the model's generalizability to unseen  categories. 
% Different from them, our work is the first one to research on the open-vocabulary VidVRD setting.
% The aim is to generalize the model to recognize the categories totally unrelated to the seen (training) ones.

% There is also a recent work~\citep{he2022towards} using prompt tuning for visual relations in images. However, their prompt representations are %
% % still statically learned 
% learned from the visual cues of static subject and object. In contrast, our RePro is to learn the compositional prompt by leveraging the motion cues in subject and object tracklets. It is expected to perform better than static prompts for detecting the dynamic visual relations in videos.

\textbf{Prompt Tuning for Open-vocabulary Visual Recognition}. \Ra{Prompt tuning~\citep{liu2021pre} has been widely adopted in both image (\eg, open-vocabulary object detection (OV-Det)~\citep{gu2021open,du2022learning,ma2022open}) and video (\eg, zero-shot video action recognition~\citep{lin2022frozen,ni2022expanding,ju2022prompting,nag2022zero}) domains.}
For OV-Det, recent works mainly focus on knowledge distillation from VLM and simply using handcrafted prompt~\citep{gu2021open,ma2022open}, or focus on  prompt representation learning for object regions~\citep{du2022learning,feng2022promptdet}.
\Ra{For video action recognition, existing works mainly use fixed prompt~\citep{nag2022zero}, conventional learnable prompt~\citep{ju2022prompting}, or prompt conditioned on the input video contexts~\citep{ni2022expanding}, and they all focus on the cross-frame attention or feature interaction.} In contrast, our RePro learns the compositional prompt by leveraging the motion cues of subject-object pairs, and \Ra{has better cross-category generalization ability to detect \emph{novel} visual relations in videos.}

\section{Method} %%%% 
% We build the Open-Vocabulary setting for VidVRD as follows. 
To build the Open-VidVRD setting, we first divide the categories of a dataset into \emph{base} and \emph{novel} splits. 
Specifically, we denote $\mathcal{C}_b^O$ and $\mathcal{C}_n^O$ as the sets of \emph{base} and \emph{novel} object categories, respectively. We use $\mathcal{C}_b^P$ and $\mathcal{C}_n^P$ to denote the sets of \emph{base} and \emph{novel} predicate categories, respectively. 
In the training stage, we use all visual relation triplet samples from $\mathcal{C}_b^O\times \mathcal{C}_b^P \times \mathcal{C}_b^O$. In the testing stage, we evaluate the model with the triplets sampled from all categories, \ie, $\mathcal{C}_b^O \cup \mathcal{C}_n^O$ and  $\mathcal{C}_b^P \cup \mathcal{C}_n^P$.
% We detect the object and relation in a naive independent manner (\eg, the independent baseline in \cite{shang2021video}) since our work mainly focus on open-vocabulary detecting ability of the model. 

\begin{figure}[t]
    \centering
    \vspace{-3ex}
    \includegraphics[width=\linewidth]{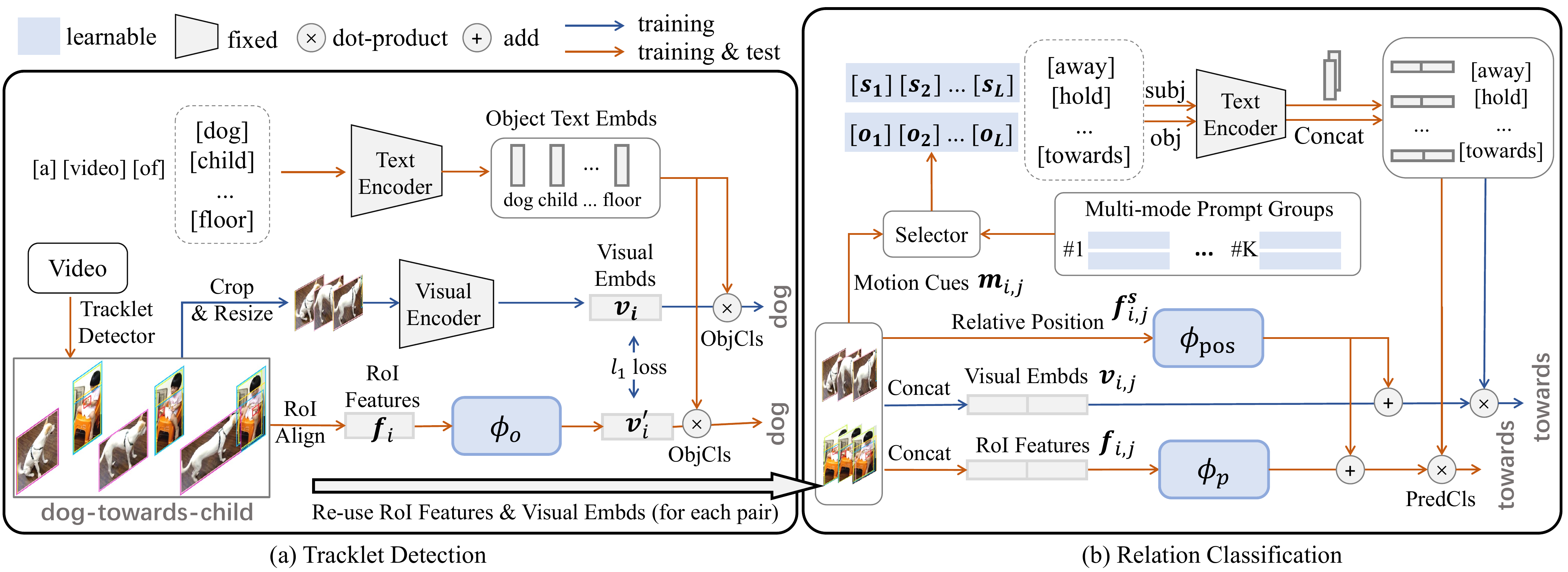}
    \vspace{-4ex}
    \caption{\Rb{The overall pipeline of RePro. The visual embeddings ($\bm{v}_i,\bm{v}_{i,j}$) are only used at training time, in which we train two V2L project modules (\ie, $\phi_o$ and $\phi_p$) to transfer the knowledge from the pre-trained VLM. At test time, only RoI features ($\bm{f}_i,\bm{f}_{i,j}$) and position features ($\bm{f}^s_{i,j}$) are used. For tracklet detection~(a), the knowledge is transferred by aligning $\bm{v}'_i$ to $\bm{v}_i$. For relation classification~(b), the knowledge is transferred by the prompt representations learned in the supervision of $\bm{v}_{i,j}$.
    % (a)~We first detect the object tracklet regions and classify them by training a V2L projector $\phi_o$. (b)~For each tracklet pair, we learn the compositional prompt based on motion cues, and classify their predicates by learning projectors $\phi_\text{pos}$ and $\phi_p$. We show only one predicate in the figure for clarity.
    }}
\label{fig_pipeline}
\vspace{-2ex}
\end{figure}

Based on this new setting, we first briefly introduce the preliminaries for open-vocabulary classification with pre-trained VLMs (Sec.~\ref{Sec_pre}). Then, we introduce the proposed Open-VidVRD method RePro, as illustrated in Figure~\ref{fig_pipeline}, in which we first extend open-vocabulary object detection methods~\citep{gu2021open,du2022learning} to open-vocabulary tracklet detection (Sec.~\ref{Sec_TrajCls}), and then perform open-vocabulary relation classification for each tracklet pair (Sec.~\ref{Sec_RelCls}).

% MoPro has two stages, as shown in Figure X. The first stage aims to learn a set of prompt representation groups with the guide of spatio-temporal motion patterns of subject-object pairs, and the second stage aims to learn a visual-to-language projection module which  aims to further utilize roi feature as training data and %%%% 这个可以移到 RelationClassification 的sebsection中。

% % enables the model get rid of the heavy pipeline of pre-trained visual encoder at the inference time.

% \gkf{we extend a ov-det method to ov-tracklet detection} %% 把这个算在我们的方法里面
% Based on these pre-extracted features, we perform object tracklet classification (Sec.~\ref{Sec_TrajCls}) and relation classification (Sec.~\ref{Sec_RelCls}) in an 
% independent manner~\cite{shang2021video}, since our work mainly focus on the open-vocabulary detecting ability of the model 
%##################Problem: 需要强调这个 independent manner 吗？？或者这一点也可以不说？

\subsection{Preliminaries: Open-Vocabulary Classification with Prompt}\label{Sec_pre}
%%%$$$$$$$$$$$$ 这里就 general的写直接用pre-train VLM 做分类。同时包括进 tracklet classification & relation classification.

\textbf{Fixed Prompt}.
Pre-trained VLMs have a strong open-vocabulary classification ability~\citep{li2022align}. They first extract text embeddings for all categories by feeding handcrafted prompt (\eg, ``a video of [CLASS]'') into the text encoder of VLM, where [CLASS] can be replaced with the class name of an arbitrary object or predicate. 
Their output text embedding $\bm{t}_c \in \mathbb{R}^d$ for each class $c$ is
\begin{equation}\label{Eq_Wc}
    \bm{t}_c = \text{VLM}_\text{txt}(\bm{W}_c), ~
    % \text{where}~ 
    \bm{W}_c=[\bm{w}_1,\ldots,\bm{w}_L,\tilde{\bm{w}}_c], 
    \forall c \in \mathcal{C}_b^O \cup \mathcal{C}_n^O ~ \text{or}~ \mathcal{C}_b^P \cup \mathcal{C}_n^P,
\end{equation}
where $\bm{W}_c$ is the prompt representation with $L$ context token vectors and the class token vector $\tilde{\bm{w}}_c$. 
% We denote the class text embeddings as $\bm{t}_c \in \mathbb{R}^d$ for each category $c$. 
% % $c \in \mathcal{C}$, where $\mathcal{C}$ can be $\mathcal{C}_b^O \cup \mathcal{C}_n^O$ for object or $\mathcal{C}_b^P \cup \mathcal{C}_n^P$ for predicate. 
Then, for each object tracklet with cropped video region, the corresponding visual embedding can be extracted by the visual encoder of VLM, denoted as \Ra{$\bm{v}_i\in \mathbb{R}^d$}. Similarly, \Ra{the visual embedding of tracklet pair ($i,j$)} can also be extracted (\eg, based on the union region), denoted as \Ra{$\bm{v}_{i,j}$}. Therefore, the $i$-th region (generally denoted as $\bm{v}_i$) can be classified by the cosine similarities w.r.t $\{\bm{t}_c\}$: %, \ie,
% \begin{equation}
%     c_i = \arg\max_{c} ~\bm{v}_i^{\rm T}\bm{t}_c / (\|\bm{v}_i\| \|\bm{t}_c\|), ~ \forall c \in \mathcal{C}_b^O \cup \mathcal{C}_n^O.
% \end{equation}  %%% TODO modify this Equation
\begin{equation}\label{Eq_argmax}
    \hat{c}_i = \arg\max_{c} \cos(\bm{v}_i,\bm{t}_c), ~ \forall c \in \mathcal{C}_b^O \cup \mathcal{C}_n^O ~ \text{or}~ \mathcal{C}_b^P \cup \mathcal{C}_n^P, ~\text{where} \cos(\bm{x},\bm{y}) = \bm{x}^{\rm T}\bm{y} / (\|\bm{x}\| \|\bm{y}\|).
\end{equation}
\textbf{Learnable Prompt}.
%%% 这里 general的写一下 prompt 是怎么学的， 后面再详细写 W_c^s  &  W_c^o; 这里只写一个 learnable prompt representation的 W_c 以及 Enco_{txt}(W_c)
Manually tuning the words in the prompt requires domain expertise and is time-consuming or not robust~\citep{radford2021learning}. The substitute method is to learn the prompt representations from the training data~\citep{zhou2022learning,zhou2022conditional}. Specifically, the context vector $\bm{w}_i$ in $\bm{W}_c$ can be set as a learnable vector while $\tilde{\bm{w}}_c$ is kept as fixed. In the training stage, samples and $\{\tilde{\bm{w}}_c\}$ are from base categories. 
% $\tilde{\bm{w}}_c$ is from the text embeddings of base classes. 
In the testing stage, the $L$ learned vectors in each $\bm{W}_c$ are fixed and then the model performs classification in the same way as Eq.~(\ref{Eq_argmax}).
% Then the prompt representations are trained with base class data.

% The prompt representation for each class $c$ can be denoted as
% \begin{equation}
%     \bm{W}_c=[\bm{w}_1,\ldots,\bm{w}_L,\tilde{\bm{w}}_c],
% \end{equation}
% where $\bm{w}_i$ is the learnable context vector and $\tilde{\bm{w}}_c$ is the fixed token for class $c$ (\ie, corresponding to the text [CLS]). In the training stage, $c$ is selected from base classes
% the open-vocabulary setting, 

% (for $c \in \mathcal{C}_b^P$ in training and for all $c$ at test time). Then the predicate text embedding $\bm{t}_c^p$ is extracted by the text encoder of VLM based on the two prompts and concatenated, \ie,
% \begin{equation}
%     \bm{t}_c^p = [\text{Enco}_\text{txt}(\bm{W}_c^s),\text{Enco}_\text{txt}(\bm{W}_c^s)],~\text{and}~\bm{t}_c^p \in \mathbb{R}^{2d}.
% \end{equation}

\subsection{Open-Vocabulary Object Tracklet Detection}\label{Sec_TrajCls}

\textbf{Tracklet Proposal Generation}.
Given a video, we first detect all the class-agnostic object tracklets using a pre-trained tracklet detector, denote as $\mathcal{T} = \{T_i\}_{i=1}^N$, as shown in Figure~\ref{fig_pipeline}(a). Specifically, each tracklet $T_i$ is characterized with \Ra{a bounding box sequence %$\bm{b}_i\in\mathbb{R}^{l_i \times 4}$ 
and the corresponding RoI Aligned~\citep{he2017mask} visual feature,} %$\bm{f}_i^0\in\mathbb{R}^{l_i \times 2048}$, where $l_i$ is the length (number of frames) of $T_i$. 
To reduce the computational overhead, \Ra{we average the RoI features of all bounding boxes (\ie, along the temporal axis of the tracklet)} following~\citep{shang2021video}, and denote it as $\bm{f}_i \in \mathbb{R}^{2048}$.

\textbf{Tracklet Classification}. Instead of directly classifying object tracklets using VLM as Eq.~(\ref{Eq_argmax}), we train a visual-to-language (V2L) projection module $\phi_o(\cdot)$ to further utilize the annotations of base classes. In particular, $\phi_o(\cdot)$ maps the RoI Aligned feature $\bm{f}_i$ of each tracklet to the same semantic space $\mathbb{R}^d$, \ie, $\Ra{\bm{v}'_i} = \phi_o(\bm{f}_i)$. 
Let $\bm{t}_c^o$ be the text embedding of object class $c \in \mathcal{C}_b^O$. The probability of tracklet $T_i$ being classified as class $c$ can be calculated as 
\begin{equation}
    p_i(c) = 
    \frac{\exp(\cos(\Ra{\bm{v}'_i},\bm{t}^o_{c})/\tau)}{\sum_{c' \in \mathcal{C}^O_b} \exp(\cos(\Ra{\bm{v}'_i},\bm{t}^o_{c'})/\tau)}, \forall c \in \mathcal{C}^O_b,
\end{equation}
where $\tau$ is a temperature parameter for softmax.

\textbf{Training Objectives}.
To train the object tracklet classification module, we assign base category labels to detected tracklets according to the IoU w.r.t ground-truth tracklets. We call those tracklets with assigned labels as positive tracklets, otherwise negative tracklets. Note that
two cases can be recognized as negative tracklets: 1) the content is background; and 2) the content contains a novel object category.
% negative tracklets might be ``true'' background, or foreground with novel object categories.
% So we do not classify them as background. 
% \gkf{mark}
%
For these negative tracklets, we follow the loss used by \cite{du2022learning}
that 
forces the prediction (from any negative tracklet) on each base class to be $1/|\mathcal{C}^O_b|$, \ie, unlike any base category. 
Therefore, the classification loss for positive and negative tracklets can be calculated as:
\begin{equation}\label{eq_trajcls}
    \mathcal{L}_\text{cls-pos} = -\frac{1}{|\mathcal{T}_p|}
    \sum_{T_i\in\mathcal{T}_p} \sum_{c\in\mathcal{C}^O_b} 
    \mathbbm{1}_{\{c=c^*_i\}} \log p_i(c), \quad
    \mathcal{L}_\text{cls-neg} = -\frac{1}{|\mathcal{T}_n|}
    \sum_{T_i\in\mathcal{T}_n} \sum_{c\in\mathcal{C}^O_b} 
    \frac{1}{|\mathcal{C}^O_b|} \log p_i(c),
\end{equation}
where $\mathcal{T}_p$ and $\mathcal{T}_n$ are the sets of positive and negative tracklets, respectively (\ie, $\mathcal{T}_p \cup \mathcal{T}_n=\mathcal{T}$), and $c^*_i$ is the ground-truth label for the $i$-th positive tracklet. We empirically found (in Sec.~\ref{Sec_TrajCls}) that using the above negative tracklet loss works better than using the loss with a unique ``background'' class~\citep{zareian2021open,gu2021open}. Besides, following~\citep{gu2021open}, \Rb{we distill the knowledge} from a pre-trained visual encoder to $\phi_o(\cdot)$ by \Ra{aligning $\bm{v}'_i$ to $\bm{v}_i$} using $l_1$ loss, \ie,
\begin{equation}\label{eq_l1}
    \Rb{\mathcal{L}_\text{distill}} = 
    % \frac{1}{N}
    (1/N)
    {\textstyle{\sum}}_{i=1}^N \|\Ra{\bm{v}'_i - \bm{v}_i}\|_1
\end{equation}
Therefore, the overall loss for object tracklet classification is $\mathcal{L}_\text{cls} = \mathcal{L}_\text{cls-pos} + \mathcal{L}_\text{cls-neg} + \lambda \Rb{\mathcal{L}_\text{distill}}$, where $\lambda$ is a hyper-parameter to weight the classification and distillation.

%%%%%%%%%%%%%%%%%%%%%%%%%%%%

% In this paper, we train a set of prompt representation groups with the guide of spatio-temporal motion patterns of subject-object pairs.

\subsection{Open-Vocabulary Visual Relation Classification}\label{Sec_RelCls}

Based on the classified object tracklets, we perform open-vocabulary relation classification for each tracklet pair, as shown in Figure~\ref{fig_pipeline}(b). 
First, we learn the prompt representations based on the pre-extracted visual embeddings of tracklet pairs, for which we introduce the \emph{compositional prompt representations} and the \emph{motion-based prompt groups}. 
Second, we utilize the pre-extracted RoI Aligned features to train a visual-to-language (V2L) projection module based on the learned prompt representations. 
Finally, for testing, we extract all predicate text embeddings and classify the predicates of each tracklet pair by using the RoI Aligned features and the trained V2L projection module.
% only use the learned prompts and the V2L projection module, and get rid of the visual encoder of VLM.

\textbf{Compositional Prompt Representations}. %% 介绍的时候稍许提一下compoistion，比如，这种方式可以叫 compositional prompt learning for predicates
% In contrast to object classes, the prompt description for predicate is not accurate. (xxx this in Introduction). For example, ``person-sit\_on-sofa'' and ``dog-sit\_on-ground'' share the same relation but different they have totally different visual context. directly model the prompt context of predicate might be suboptimal because the VLM are mainly pre-trained with noun data, not action related data.  %%% 这一部分应该放到 introduction 中
% The compositional prompt learns the prompt context for subject and object separately.
The compositional prompt consists learnable prompt representations $\bm{S}_c$ and $\bm{O}_c$ (of predicate class $c$) for subject and object, respectively:
\begin{equation}\label{Eq_prompt}
    \bm{S}_c=[\bm{s}_1,\ldots,\bm{s}_L,\tilde{\bm{w}}_c], \quad \bm{O}_c=[\bm{o}_1,\ldots,\bm{o}_L,\tilde{\bm{w}}_c], %\quad \forall c \in \mathcal{C}_b^P,
\end{equation}
\Rb{where $\bm{s}_i$ and $\bm{o}_i$ are the learnable context vectors} and $\tilde{\bm{w}}_c$ is the fixed class token for predicate $c$ (for $c \in \mathcal{C}_b^P$ in training phase and for all $c$ in testing phase). 
% Then, the predicate text embedding $\bm{t}_c^p$ is extracted by the text encoder of VLM based on the two prompts and concatenated, \ie,
Then, the predicate text embedding $\bm{t}_c^p$ is generated by concatenating the two outputs of VLM given two prompts (respectively) as inputs, \ie,
% by the text encoder of VLM based on the two prompts and concatenated, \ie,
\begin{equation}
    \bm{t}_c^p = [\text{VLM}_\text{txt}(\bm{S}_c),\text{VLM}_\text{txt}(\bm{O}_c)],~\text{and}~\bm{t}_c^p \in \mathbb{R}^{2d}.
\end{equation}

\textbf{Motion-based Prompt Groups}. % motion based multi-mode prompts
% Furthermore, we
% We make the prompt contexts vary from
We vary the prompt contexts based on the motion cues, \ie, the relative spatio-temporal motion patterns, between each pair of subject and object.
% , termed as motion cues (motion patterns).
In specific, we take the generalized IoU~\cite{Rezatofighi_2018_CVPR} (\ie, GIoU) as the metric to calculate the motion patterns. For each tracklet pair ⟨$T_i,T_j$⟩, % we use a vector to
we use a vector to represent a motion pattern:
% $\bm{m}_{i,j} = [g^s_{i,j},g^e_{i,j},g^s_{i,j}-g^e_{i,j}]$, 
\begin{equation}\label{Eq_motion}
    \bm{m}_{i,j} = \sign([G^s_{i,j} - \gamma,G^e_{i,j} - \gamma, G^e_{i,j} - G^s_{i,j}]), ~~ \text{and} ~~\bm{m}_{i,j} \in \{+,-\}^{3},
    %\quad \text{where} ~~ g^*_{i,j} = \text{GIoU}(T_i,T_j).
\end{equation}
% \begin{equation}
%     \bm{m}_{i,j} = [GIoU_s(i,j) > \gamma,g^e_{i,j} > \gamma, g^e_{i,j} - g^s_{i,j} > 0], ~~ \text{and} ~~\bm{m}_{i,j} \in \{0,1\}^{3},
%     %\quad \text{where} ~~ g^*_{i,j} = \text{GIoU}(T_i,T_j).
% \end{equation}
where $G^s_{i,j}, G^e_{i,j}$ are the GIoU between subject-object for the start and end bounding boxes of their temporal intersection, respectively, and $\gamma$ is a threshold for GIoU. This definition considers %the motion pattern from 
two perspectives: 1) whether the two tracklets are near or far (\ie, the first two terms of Eq.~(\ref{Eq_motion})), and 2) whether they move toward or away to each other (\ie, the third term of Eq.~(\ref{Eq_motion}). 
Overall, we have 6 motion patterns (cf. Sec.~\ref{App_motion} for more details) and build 6 prompt groups correspondingly. Each group consists its own compositional prompt representations $\bm{S}_c$ and $\bm{O}_c$ as defined in Eq.~(\ref{Eq_prompt}). 

% At the training stage, we sample the tracklet pair for each motion pattern, and calculate the training loss for each prompt group as Eq.~(\ref{Eq_predcls}), the total loss is averaged across all groups. \gkf{At the test time, we calculate the motion pattern of the tracklet pair, and then select the corresponding learned prompt representation (not ensemble) to generate the class text embeddings. Finally, we classify its visual relations as Eq.~(\ref{Eq_sigmoid}). (this is the test for stage-1)}

It's worth noting that we aim to build a framework for learning
motion-based multi-mode prompts.
The used GIoU-based approach (in our framework) is \Rb{a simple and intuitive way to calculate} motion cues.
% just one simple way to calculate the motion pattern. 
This approach is not perfect, \eg, it is poor to capture the motion pattern of tracklets moving back and forth. 
We leave other fancier (motion capturing) approaches as future work.
% The comprehensive analyses and other fancy metrics of motion pattern are left to future work.

% 不要说naive way，就说我们用的是最简单的一种实验，更加comprehensive或者complex的就留给future work
% motion pattern of subject-object pair. The motion pattern indicates the relative position of the subject and object tracklets. In our work, 

% \textbf{Training Objectives}.  % %%% 两个stage 的loss部分是一样的，应该先给一个high level 的叙述，讲我们有两个stage，然后再讲每个stage的loss

\textbf{Training Objectives}. 
Based on the above definition, we train the prompt representations with visual embeddings and relative position features. For simplicity, we show the training process of a single group (and in the end, we derive the final loss by averaging the losses across all groups). For each tracklet pair ⟨$T_i, T_j$⟩, we first calculate its motion cue, select the corresponding prompt group, and extract the class text embeddings $\bm{t}_c^p$ for each predicate class $c \in \mathcal{C}_b^P$. 
Then, we take the pre-extracted visual embeddings $\bm{v}_i$ and $\bm{v}_j$, and concatenate them as the pair's visual embedding $\Ra{\bm{v}_{i,j} = [\bm{v}_i,\bm{v}_j] \in \mathbb{R}^{2d}}$. 
Following~\cite{shang2021video}, we additionally compute the relative position feature between bounding boxes of $T_i$ and $T_j$, 
% and concatenate the results\qianru{what do you mean by "results" -- bboxes or positions?} of start and end bounding boxes, 
denoted as $\bm{f}^{s}_{i,j} \in \mathbb{R}^{12}$ (cf. Sec.~\ref{App_motion} in the Appendix for more details). 
The predicted probability of predicate class $c$ in this tracklet pair is thus
\begin{equation}
    \label{Eq_sigmoid}
    % \bm{p}_{i,j} = [p_{i,j}(c)],\quad \text{where}~ 
    p_{i,j}(c) = \text{Sigmoid}(\cos(\Ra{\bm{v}_{i,j}}+ \phi_{\text{pos}}(\bm{f}^s_{i,j}), \bm{t}_c^p)), ~ \forall c \in \mathcal{C}_b^P.
\end{equation}
where $\phi_{\text{pos}}$ projects $\bm{f}_{i,j}^s$ to the same dimension as $\Ra{\bm{v}_{i,j}}$. Going through all \emph{base} classes (in $\mathcal{C}_b^P$), the probability vector of tracklet pair ⟨$T_i,T_j$⟩ is generated and can be denoted as $\bm{p}_{i,j}$, \ie, each dimension is calculated by Eq.~(\ref{Eq_sigmoid}). 
In the training time, we assign the predicate labels according to the IoU for each tracklet pair w.r.t the ground-truth tracklet pair by following \cite{shang2021video}. We denote the sets of positive and negative tracklet pairs as $\mathcal{P}_p$ and $\mathcal{P}_n$, respectively. Due to the multi-label setting of VidVRD, we use binary cross-entropy loss for relation classification. The ground truth for positive tracklet pair in $\mathcal{P}_p$ is a binary vector of dimension $|\mathcal{C}_b^P|$, denoted as $\bm{p}^*_{i,j}$. 
% Note that we do not define a background embedding $\bm{t}^p_\text{bg}$ on purpose, and 
For those negative tracklet pairs in $\mathcal{P}_n$, we optimize the probability of each \emph{base} class to zero, \ie, the ground truth is an all-zero vector. The classification loss is thus calculated as
% \begin{equation}\label{Eq_predcls}
%     \mathcal{L}_\text{pred-cls} = \frac{1}{|\mathcal{P}_p|} {\textstyle \sum}_{(T_i,T_j)\in\mathcal{P}_p} \text{BCE}(\bm{p}_{i,j},\bm{p}^*_{i,j}) + \frac{1}{|\mathcal{P}_n|}{\textstyle \sum}_{(T_i,T_j)\in\mathcal{P}_n} \text{BCE}(\bm{p}_{i,j},\bm{0}).
% \end{equation}
\begin{equation}\label{Eq_predcls}
    \mathcal{L}_\text{pred-cls} = (1/|\mathcal{P}_p|) {\textstyle \sum}_{(T_i,T_j)\in\mathcal{P}_p} \text{BCE}(\bm{p}_{i,j},\bm{p}^*_{i,j}) + (1/|\mathcal{P}_n|){\textstyle \sum}_{(T_i,T_j)\in\mathcal{P}_n} \text{BCE}(\bm{p}_{i,j},\bm{0}).
\end{equation}

\textbf{Training V2L Projection Module}. Once the prompt representations are learned, we train a visual-to-language (V2L) projection module to use RoI Aligned features $\{\bm{f}_{i}\}$ as training data, and get rid of VLM's visual encoder at inference time. 
Given the learned prompt representations, we pre-extract the predicate class text embeddings (denoted as $\{\tilde{\bm{t}}_c^p\}$) for each prompt group and fix them. 
% For simplicity, we only show the training process of one group. 
Formally, for each tracklet pair ⟨$T_i,T_j$⟩, we concatenate their RoI features as $\Ra{\bm{f}_{i,j} = [\bm{f}_i,\bm{f}_j] \in \mathbb{R}^{4096}}$. Then, we use a V2L projection module $\phi_p$ to project it to the same dimension as text embeddings. Similar to Eq.~(\ref{Eq_sigmoid}), the probability of predicate class $c$ is predicted as
\begin{equation}
    p_{i,j}(c) = \text{Sigmoid}(\cos(\phi_p(\Ra{\bm{f}_{i,j}}) + \phi_\text{pos}(\bm{f}^{s}_{i,j}), \tilde{\bm{t}}_c^p)), ~ \forall c \in \mathcal{C}_b^P.
\end{equation}
where $\phi_\text{pos}$ is the learned spatio-temporal projection layer and is fixed. Then, we apply the same loss as defined in Eq.~(\ref{Eq_predcls}), and compute the final total loss by averaging across all groups. 

\textbf{Discussions}. Intuitively, we can train the prompt representations together with the V2L projection module $\phi_p$, and use the $l_1$ loss to align $\phi_p(\Ra{\bm{f}_{i,j}})$ to $\Ra{\bm{v}_{i,j}}$, \ie, distill the knowledge from the pre-trained visual encoder to the V2L module.  We name this variant as RePro$^\dagger$.
We justify that our RePro works better than RePro$^\dagger$ due to: 1) directly using the teacher (\ie, $\Ra{\bm{v}_{i,j}}$) to train the prompt is intuitively better than using student (\ie, projected visual embedding), and 2) the distillation makes the V2L module focus too much on the static visual alignment, rather than the dynamic relation information learned in the prompt. In experiments, we empirically show the superiority of RePro over RePro$^\dagger$.
% and the experiment results in Sec.~\ref{Sec_OpenVocRelCls} have shown the superiority of the proposed training scheme over the above unified training scheme of RePro$^\dagger$. 

%$$$$$$$$$$$$$$ 这里就不要写two-stage， 不要提到two-stage 这个名字，这里就是说我们要先训练一个prompt，然后用训练好的prompt 来训练V2L，当然也可以 加蒸馏 一起训练。但是这样的话有两个缺点1）V2L只会focus on visual alignement，没有学到prompt中的relation信息。2）在学prompt的时候，需要用最准确的visual embedding引导，既然要蒸馏，那不如直接用teacher。 <--- 把review 引导到对比这两件事情上来。而不是让他们去想stage-1 和 stage-2的对比。 因为我们没有单独的stage-1的performance的实验。。

%%%% 不要专门起个 second-stage的名字，因为我们没有实验和第一个stage 比，也不要用这个小标题。直接就是写在 Training Objectives 里面

\section{Experiments}\label{sec_exps}
\subsection{Datasets and Evaluation Metrics}
\textbf{Datasets}. We evaluated our method on the VidVRD~\citep{shang2017video} and VidOR~\citep{shang2019annotating} benchmarks: 1) VidVRD consists of 1,000 videos, and covers 35 object categories and 132 predicate categories. We used official splits: 800 videos for training and 200 videos for testing. 2) VidOR consists of 10,000 videos, which covers 80 object categories and 50 predicate categories. We used official splits: 7,000 videos for training, 835 videos for validation, and 2,165 videos for testing. Since the annotations of VidOR-test are not released, we only evaluated models on validation set.

\textbf{Evaluation Settings}. 
To build the open-VidVRD setting, we manually split \emph{base} and \emph{novel} categories by selecting the common object and predicate categories as the \emph{base} split, and selecting the rare ones as the \emph{novel} split. The detailed splits are given in Sec.~\ref{App_split} of the Appendix. 
% Specifically, for VidVRD, the number of base/novle categories is 25/10 for object, and 71/61 for predicate. For VidOR, the number of base/novel categories is 50/30 for object and 30/20 for predicate (refer to the Appendix for detailed categories). 
We trained the model on the triplet samples of both \emph{base} object and predicate categories in the training set. During testing, we evaluated the model on two settings: 1) \textbf{Novel-split}: triplet samples with all object categories and \emph{novel} predicate categories, 
% cates, 
and 2) \textbf{All-splits}: triplet samples with all object and predicate categories, in the testing set of VidVRD (or the validation set of VidOR).

\textbf{Metrics}. 
We follow three standard evaluation tasks in scene graph generation~\citep{zellers2018scenegraphs}: scene graph detection (SGDet), scene graph classification (SGCls), and predicate classification (PredCls). We apply these metrics to VidVRD: a detected triplet is considered to be correct if there is the same triplet tagged in the ground truth, and both subject and object tracklets have a sufficient volume IoU (\eg, 0.5) with the ground truth. %We use mAP and Recall@K (R@K, K=50,100) as evaluation metrics.
Following the standard setting~\citep{shang2017video}, we use mAP and Recall@K (R@K, K=50,100) as evaluation metrics.

\subsection{Implementation Details}
\textbf{Tracklet Detector \& Pre-trained VLM}. We used the Faster-RCNN~\citep{ren2015faster}-based VinVL model~\citep{zhang2021vinvl} to detect frame-level object bounding boxes and extracted corresponding RoI Alinged features, and then adopted Seq-NMS~\citep{han2016seq} to generate class-agnostic object tracklets. The VinVL model was trained on out-of-domain image data without seeing any VidVRD data. For pre-trained VLM, we used ALPro~\citep{li2022align}, which was pre-trained on a wide range of video-language data, and learned the fine-grained visual region to text entity alignment. %\Rb{We used the same object tracklets for all of our experiments.}

% \textbf{Relation Detection Details}. More details and hyperparameter settings are left in Sec.~\ref{app_hyperparameters}.% of the Appendix.

% Relation Detection Details

\textbf{Relation Detection Details}. Following the popular segment-based methods~\citep{qian2019video,shang2017video,shang2021video}, we first detected visual relations in short video segments, and then adopted greedy relation association algorithm~\citep{shang2017video} to merge the same relation triplets. The detailed hyperparameter settings are left in Sec.~\ref{app_hyperparameters} of the Appendix.
% Refer to Sec.~\ref{app_hyperparameters} for hyperparameter settings.

% \textbf{Hyperparameters}. The detailed hyperparameter settings are left in Sec.~\ref{app_hyperparameters} of the Appendix.

\subsection{Evaluate Open-vocabulary Object Tracklet Detection}
We evaluated the tracklet detection part of RePro on \emph{novel} object categories, as shown in Table~\ref{table:TrajCls}. 

\Rc{
\textbf{Comparison to ALPro}. A straightforward baseline to achieve open-vocabulary tracklet detection is directly applying the pre-trained VLM (ALPro in our case) by inputting the tracklet regions into its visual encoder to perform classification, as in Eq.~(\ref{Eq_argmax}). However, this has a significant computational overhead due to the heavy pipeline of ALPro's visual encoder. In contrast, our RePro requires much less computational cost, since we only use one projection layer (\ie, $\phi_o$). We thus compare our RePro with the above ALPro baseline. The results in row~\textbf{\#4} show that RePro can achieve comparable performances on both datasets, with the projection layer $\phi_o$.%We further demonstrate the effectiveness of our design.
% of RePro's tracklet detection part.
}

\textbf{Negative Tracklet Classification}. \Rc{How to model the negative sample is a key challenge as widely discussed in many open-vocabulary object detection works. There are usually two approaches: 1) using a unique background embedding (BG-Embd) in addition to the class text embeddings~\citep{zareian2021open,gu2021open}, and 2) only using the class text embeddings, and computing the loss of negative sample as $\mathcal{L}_{\text{cls-neg}}$ in Eq.~(\ref{eq_trajcls})~\citep{gu2021open}.} By comparing rows~\textbf{\#3} and \textbf{\#4} of Table~\ref{table:TrajCls}, we find that without using background embedding (\ie, as $\mathcal{L}_{\text{cls-neg}}$) achieves better recall, and outperforms the other by a large margin, especially on the more challenging VidOR benchmark. This is because the tracklets recognized as negative may be due to the fact that they contain novel objects (rather than backgrounds), and aligning their embeddings (\ie, different novel class embeddings) to a unique background embedding hurts the model's recognition ability on novel objects.

\begin{table}[t]
    \vspace{-3ex}
    \caption{Performance (\%) of tracklet detection on objects with novel categories.}
    \label{table:TrajCls}
    \vspace{-2ex}
    \centering
    % \begin{tabular}{lccc}
    % \toprule
    % Supervision & AR$_r@100$ & AR$_r@300$ & AR$_r@1000$ \\
    % \midrule
    % base & 39.3 & 48.3 & 55.6 \\
    % base + novel & 41.1 & 50.9 & 57.0 \\
    % \bottomrule
    % \end{tabular}
    \begin{tabular}{lccccccc}
        \hline
        \multicolumn{2}{c}{\multirow{2}{*}{Methods}} & \multirow{2}{*}{Distillation} & \multirow{2}{*}{BG-Embd} & \multicolumn{2}{c}{VidVRD-test} & \multicolumn{2}{c}{VidOR-val}   \\
        \multicolumn{2}{c}{}                         &                               &                           & R@5            & R@10           & R@5            & R@10           \\ \hline
        \multicolumn{2}{l}{ALPro~\citep{li2022align}}                    & -                             & -                         & 41.38          & 53.81          & 34.26          & 41.72          \\ \hline
        \multirow{4}{*}{RePro}         & \#1        & $\times$       & $\times$                         & 2.17           & 3.71           & 2.33           & 3.48           \\
                                        & \#2        & $\times$                             & \checkmark                     & 32.43          & 33.36          & 7.58           & 12.37          \\
                                        & \#3        & \checkmark                         & \checkmark                     & 43.84          & \textbf{53.00} & 12.61          & 16.85          \\
                                        & \#4        & \checkmark                         & $\times$                         & \textbf{46.34} & 50.42          & \textbf{31.62} & \textbf{37.08} \\ \hline
        \end{tabular}
    \vspace{-2ex}
\end{table}

\begin{table}[h]
    \vspace{-3ex}
    \caption{Performance (\%) comparision to conventional methods on VidVRD-test. Relation Tagging (RelTag) only considers the precision of relation triplets and ignores the localization of tracklets.}
    \label{table:VidVRD-conventional-sota}
    \vspace{-2ex}
    \centering
    % \addtolength{\tabcolsep}{-1pt}
    \begin{tabular}{lccccccc}
        \hline
        \multirow{2}{*}{Methods} & \multirow{2}{*}{Training Data} & \multicolumn{3}{c}{SGDet} & \multicolumn{3}{c}{RelTag}                                                        \\
&           & mAP  & R@50 & R@100& P@1  & P@5  & P@10                      \\ \hline
        
        \cite{su2020video}    & base+novel       & 19.03& 9.53 & 10.38& 57.50& 41.40& 29.45                     \\
        \cite{liu2020beyond}   & base+novel       & 18.38& 11.21& 13.69& 60.00& 43.10& {32.24}            \\
        \cite{li2021interventional}  & base+novel   & \textbf{22.97} & {12.40} & {14.46} & \textbf{68.83} & \textbf{49.87} & \textbf{35.57} \\
        \cite{gao2022classification}  & base+novel    & 17.67        & 9.63 & 11.29& 56.00& 43.80& \textbf{32.85}                     \\
        RePro (Ours)      & base      & 21.33& \textbf{12.92}& \textbf{15.94}& {59.00}& {41.09}& 28.87                     \\
        \Rd{RePro (Ours)}      & \Rd{base+novel}      & \Rd{\textbf{25.55}} & \Rd{\textbf{13.83}}& \Rd{\textbf{17.33}} & \Rd{\textbf{62.50}}& \Rd{\textbf{45.80}}& \Rd{{32.05}}           \\
        \hline
        \end{tabular}
    \vspace{-1ex}
\end{table}

% 21.33	12.92	15.94	59.00	41.09	28.87
% 25.55	13.83	17.33	62.50	45.80	32.05

%  base+novel   | **25.02** | **13.81** | **17.22** | **67.50** | **50.10** | **36.50** |

% \rotatebox[origin=c]{90}{Novel}
\begin{table}[h]
    % \vspace{-1ex}
    \caption{Performance (\%) comparision of Open-VidVRD methods on VidVRD-test.}
    \label{table:VidVRD-OpenVoc-sota}
    \vspace{-2ex}
    \centering
    \addtolength{\tabcolsep}{-2.5pt}
    \begin{tabular}{llccccccccc}
        \hline
        \multirow{2}{*}{Split} & \multirow{2}{*}{Methods} & \multicolumn{3}{c}{SGDet}                        & \multicolumn{3}{c}{SGCls}                        & \multicolumn{3}{c}{PredCls}                      \\
                               &                          & mAP            & R@50           & R@100          & mAP            & R@50           & R@100          & mAP            & R@50           & R@100          \\ \hline
        \multirow{4}{*}{Novel} & ALPro                    & 1.05	& 3.14	& 4.62	& 3.69	& 7.27	& 8.92	& 4.09	& 9.42	& 10.41          \\
                               & VidVRD-II                & 3.57	& 8.59	& 12.39	& 5.70	& 13.22	& 18.34	& 7.35	& 18.84	& 26.44          \\
                               & RePro$^\dagger$          & 2.56	& 8.26	& 11.73	& 8.63	& 15.04	& 18.84	& 9.34	& 18.67	& 24.13          \\
                               & RePro                    & \textbf{6.10}	& \textbf{13.38}	& \textbf{16.52}	& \textbf{10.32}	& \textbf{19.17}	& \textbf{25.28}	& \textbf{12.74}	& \textbf{25.12}	& \textbf{33.88} \\ \hline
        \multirow{4}{*}{All}   & ALPro                    & 3.20	& 2.62	& 3.18	& 3.92	& 3.88	& 4.75	& 4.97	& 4.50	& 5.79           \\
                               & VidVRD-II                & 12.74	& 9.90	& 12.59	& 17.26	& 14.93	& 19.68	& 19.73	& 18.17	& 24.90          \\
                               & RePro$^\dagger$          & 16.21	& 11.14	& 14.56	& 22.37	& 16.83	& 21.71	& 25.43	& 21.36	& 28.04          \\
                               & RePro                    & \textbf{21.33}	& \textbf{12.92}	& \textbf{15.94}	& \textbf{30.15}	& \textbf{19.75}	& \textbf{25.00}	& \textbf{34.90}	& \textbf{25.50}	& \textbf{32.49} \\ \hline
        \end{tabular}

    \vspace{-2ex}
\end{table}

% SGDet			SGCls			PredCls		
% mAP	R@50	R@100	mAP	R@50	R@100	mAP	R@50	R@100
% & 1.05	& 3.14	& 4.62	& 3.69	& 7.27	& 8.92	& 4.09	& 9.42	& 10.41
% & 3.57	& 8.59	& 12.39	& 5.70	& 13.22	& 18.34	& 7.35	& 18.84	& 26.44
% & 2.56	& 8.26	& 11.73	& 8.63	& 15.04	& 18.84	& 9.34	& 18.67	& 24.13
% & 6.10	& 13.38	& 16.52	& 10.32	& 19.17	& 25.28	& 12.74	& 25.12	& 33.88
% & 3.20	& 2.62	& 3.18	& 3.92	& 3.88	& 4.75	& 4.97	& 4.50	& 5.79
% & 12.74	& 9.90	& 12.59	& 17.26	& 14.93	& 19.68	& 19.73	& 18.17	& 24.90
% & 16.21	& 11.14	& 14.56	& 22.37	& 16.83	& 21.71	& 25.43	& 21.36	& 28.04
% & 21.33	& 12.92	& 15.94	& 30.15	& 19.75	& 25.00	& 34.90	& 25.50	& 32.49

\textbf{Distillation}. We verified the effectiveness of visual distillation (\ie, Eq.~(\ref{eq_l1})) by comparing rows~\textbf{\#2} and \textbf{\#3} of Table~\ref{table:TrajCls}. Obviously, the distillation helps RePro improve the detection recall by a large margin, especially for the more challenging VidOR benchmark. For row \textbf{\#1}, we can observe that 
\Rc{computing the negative tracklet classification loss as $\mathcal{L}_{\text{cls-neg}}$ without distillation has extremely low performance. This is because forcing the classification probability of negative tracklet to be $1/|\mathcal{C}_b|$ (\ie, by $\mathcal{L}_{\text{cls-neg}}$) and without the guidance from the teacher (\ie, without distillation) make the model has poor generalize ability to novel categories.}

\subsection{Evaluate Open-vocabulary Relation Classification}\label{Sec_OpenVocRelCls}
\Ra{The relation classification part of our RePro was trained separately by keeping the results of tracklet detection fixed. All of our experiments for relation classification used the same tracklet detection results (which is row~\textbf{\#4} in Table~\ref{table:TrajCls}).}

\textbf{Comparison to Conventional VidVRD SOTA Methods}. We compared our RePro with several SOTA methods in the conventional VidVRD setting, and showed the results in Table~\ref{table:VidVRD-conventional-sota}. \Ra{The object tracklets and features used in SOTA methods are not uniform since VidVRD is a very challenging task (see Sec.~\ref{app_sota} for details).} We can observe that even when our RePro is trained with only base category samples (while others are with both base and novel category samples), our performance on SGDet tasks is comparable to others'. \Rd{When trained with both base and novel category samples, our RePro outperforms all other SOTA methods in all SGDet tasks and most RelTag tasks.}
% when trained and tested on base category samples\qianru{please check}, our RePro outperforms SOTA methods on the SGDet task, and has comparable performance on the Relation Tagging.

\textbf{Comparisons in the Setting of Open-VidVRD}. 
We compared the model performances in the setting of Open-VidVRD and showed results in Table~\ref{table:VidVRD-OpenVoc-sota}. \Rb{Since our RePro is the first Open-VidVRD method, we compared it to ALPro (implemented as Eq.~(\ref{Eq_argmax})). We also re-implemented the SOTA method VidVRD-II~\citep{shang2021video} and trained it on base category samples. We replaced its classifier with text embeddings extracted by ALPro's text encoder. For both ALPro and VidVRD-II, we used a fixed (handcrafted) prompt ``a video of relation [CLASS]". In addition, we reported the results of RePro's intuitive variant RePro$^\dagger$ as mentioned in the ``\textbf{Discussion}'' of Sec.~\ref{Sec_RelCls}.}

\Rb{From the results in Table~\ref{table:VidVRD-OpenVoc-sota}, we can observe that our RePro outperforms ALPro, VidVRD-II and RePro$^\dagger$ by a large margin on both Novel-split and All-splits. By comparing RePro to ALPro, we show that, unlike that in tracklet classification, directly applying pre-trained VLM to relation classification is sub-optimal and achieves poor performance. By comparing RePro to VidVRD-II, we demonstrate the superiority of our prompt tuning framework over the fixed prompt design. By comparing RePro to RePro$^\dagger$, we validate the effectiveness of our training scheme for RePro.}

\begin{table}[h]
    \vspace{-3ex}
    \caption{Ablations (\%) for RePro with different prompt design in VidVRD-test, where \textbf{C} stands for Compositional, and \textbf{M} stands for Motion cues. \textbf{Ens}: ensemble all the learned prompts by averaging their representations. \textbf{Rand}: reandomly select a prompt without considering motion cues.}
    \label{table:VidVRD-ablation}
    \vspace{-2ex}
    \centering
    \addtolength{\tabcolsep}{-2pt}
    
\begin{tabular}{ccccccccccccc}
    \hline
    \multirow{2}{*}{} & \multirow{2}{*}{} & \multirow{2}{*}{C} & \multirow{2}{*}{M} & \multicolumn{3}{c}{SGDet}      & \multicolumn{3}{c}{SGCls}      & \multicolumn{3}{c}{PredCls}    \\
  & &      &       & mAP     & R@50    & R@100   & mAP     & R@50    & R@100   & mAP     & R@50    & R@100   \\ \hline
    \multirow{5}{*}{\rotatebox[origin=c]{90}{Novel-split}}  & \#1   & $\times$    & $\times$     & 3.50	& 9.91	& 13.88	& 7.21	& 14.54	& 19.83	& 8.63	& 20.33	& 27.43   \\
                                                            & \#2   & \checkmark  & $\times$     & 5.57	& 11.40	& 14.87	& 10.31	& 16.52	& 21.81	& 11.83	& 22.31	& 30.90   \\
                                                            & \#3   & \checkmark  & Ens          & 6.24	& 11.57	& 15.20	& 10.77	& 16.03	& 21.98	& 12.36	& 21.32	& 29.91    \\
                                                            & \#4   & \checkmark  & Rand         & \textbf{7.14}	& 11.90	& 14.87	& \textbf{10.85}	& 16.52	& 23.30	& 12.42	& 22.64	& 30.90   \\
                                                            & \#5   & \checkmark  & \checkmark   & 6.10	& \textbf{13.38}	& \textbf{16.52}	& 10.32	& \textbf{19.17}	& \textbf{25.28}	& \textbf{12.74}	& \textbf{25.12}	& \textbf{33.88} \\ \hline
    \multirow{5}{*}{\rotatebox[origin=c]{90}{All-splits}}   & \#1   & $\times$    & $\times$     & 19.73& 12.26	& 15.36	& 26.80	& 18.24	& 23.06	& 30.80	& 23.70	& 30.42   \\
                                                            & \#2   & \checkmark  & $\times$     & 18.47& 11.95	& 15.28	& 25.52	& 18.13	& 23.12	& 29.45	& 23.39	& 30.17   \\
                                                            & \#3   & \checkmark  & Ens          & 20.15& 12.38	& 15.61	& 27.93	& 18.61	& 23.55	& 31.68	& 23.61	& 30.29   \\
                                                            & \#4   & \checkmark  & Rand         & \textbf{21.72}& 12.71	& 15.78	& 29.15	& 19.15	& 24.13	& 33.11	& 24.38	& 31.49   \\
                                                            & \#5   & \checkmark  & \checkmark   & 21.33& \textbf{12.92}	& \textbf{15.94}	& \textbf{30.15}	& \textbf{19.75}	& \textbf{25.00}	& \textbf{34.90}	& \textbf{25.50}	& \textbf{32.49} \\ \hline
    \end{tabular}

    \vspace{-3ex}
\end{table}

\subsection{Ablation Studies}\label{sec_ablation}
We conducted careful ablation studies as shown in Table~\ref{table:VidVRD-ablation}. \Ra{Since the compositional and motion-based prompt design is one of our main contributions, we conducted ablations w/o either of them (rows~\textbf{\#1, \#2} and \textbf{\#5}). To further show the effectiveness of our motion pattern design, we designed two variants, \ie, rows~\textbf{\#3} (Ens) and \textbf{\#4} (Rand). Their detailed settings are enumerated as follows:} 
\textbf{\#1}: It learns a single prompt representation $\bm{W}_c$ as in Eq.~(\ref{Eq_Wc}). The obtained predicate text embedding has the half dimensions of $\bm{t}_c^p$ in Eq.~(\ref{Eq_prompt}). 
% \textbf{\#1}: training with neither compositional prompt nor motion cues. Specifically, we learned a single prompt representation $\bm{W}_c$ as in Eq.~(\ref{Eq_Wc}). The obtained predicate text embedding has the half dimensions of the $\bm{t}_c^p$ in Eq.~(\ref{Eq_prompt}). 
%
So we calculated the visual embeddings of a tracklet pair as 
% \footnote{To avoid introducing extra parameters (and also avoid over-fitting to base categories), we do not use the concatenated $\Ra{\bm{v}_{i,j}}$ and train an MLP to reduce its dimension to $d$.} 
$\Ra{\bm{v}_{i,j} = \bm{v}_i - \bm{v}_j}$ (different from the concatenated vector $\Ra{\bm{v}_{i,j}}$ in Eq.~(\ref{Eq_sigmoid})). 
\textbf{\#2}: Training with compositional prompt but without motion cues. \textbf{\#3} \& \textbf{\#4}: Training with compositional prompt, but the prompt is randomly selected from the 6 groups without considering motion cues. For testing, the prompts are ensembled by  averaging (\textbf{Ens}) or randomly selected (\textbf{Rand}).
% training as in \textbf{\#3}, but the prompt is randomly selected at testing. 
\textbf{\#5}: The proposed RePro.%, where the prompt is chosen according to the motion cues.% of tracklet pairs.
% in corresponding motion group.%\qianru{please check}. 

\begin{wraptable}{r}{45ex}
    % \vspace{-3ex}
    % \begin{table}[h]
        \vspace{-2ex}
        \caption{Ablations (\%) on VidOR-val.}
        \label{table:VidOR-val2}
        % \vspace{-1ex}
        \centering
        \addtolength{\tabcolsep}{-2.5pt}
        \begin{tabular}{ccccccc}
            \hline
            \multirow{2}{*}{} & \multirow{2}{*}{C} & \multirow{2}{*}{M} & \multicolumn{2}{c}{SGCls}       & \multicolumn{2}{c}{PredCls}     \\
                                  &                    &                    & R@50           & R@100          & R@50           & R@100          \\ \hline
            \multirow{5}{*}{\rotatebox[origin=c]{90}{Novel-split}} & $\times$                  & $\times$                  & 0.86           & 0.86           & 2.30           & 2.88           \\
                                  & \checkmark                 & $\times$                  & 1.72           & 2.59           & 6.62           & 8.06           \\
                                  & \checkmark                 & Ens                & \textbf{2.30}  & \textbf{2.59}  & 7.20           & 8.35           \\
                                  & \checkmark                 & Rand               & 2.01           & 2.30           & 5.76           & 7.20           \\
                                  & \checkmark                 & \checkmark                 & 2.01           & 2.30           & \textbf{7.20}  & \textbf{8.35}  \\ \hline
            \multirow{5}{*}{\rotatebox[origin=c]{90}{All-splits}}   & $\times$                  & $\times$                  & 9.49           & 12.85          & 25.62          & 34.83          \\
                                  & \checkmark                 & $\times$                  & \textbf{10.06} & \textbf{13.40} & 27.00          & \textbf{36.73} \\
                                  & \checkmark                 & Ens                & 9.49           & 12.68          & 25.66          & 35.16          \\
                                  & \checkmark                 & Rand               & 10.03          & 13.13          & 26.94          & 36.48          \\
                                  & \checkmark                 & \checkmark                 & 10.03          & 12.91          & \textbf{27.11} & 35.76          \\ \hline
            \end{tabular}
    \vspace{-2ex}
    % \end{table}
    
    % \vspace{-2ex}
\end{wraptable}

\textbf{Compositional Prompt}.
By comparing the results in \textbf{\#1} and \textbf{\#2}, we can observe that the compositional prompt can effectively improve the performance on Novel-split. Meanwhile, the improvement on All-splits is not significant. We conjecture that the base relations, the majority of All-splits, require less compositional semantic contexts for prompt learning.

\textbf{Motion-based Prompt Groups}.
By comparing our RePro (\textbf{\#5})~vs.~\textbf{\#2}, we can observe that with the help of motion cues, our RePro achieves significant improvements of recall on all tasks, and also achieves considerable improvements in mAP on most tasks. We can see that the improvement on Novel-split is more significant than that on All-splits,  
\eg, 14.87\%$\rightarrow$16.52\%~vs.~15.28\%$\rightarrow$15.94\% on R@100 of SGDet, 
showing that the motion-based prompt has a better generalizability for detecting novel relations. 
Besides, if comparing RePro (\textbf{\#5}) to \textbf{Ens} (\textbf{\#3}), we can see that RePro outperforms \textbf{Ens} in Novel-split on most tasks, and achieves considerable improvements for All-splits on all tasks. Compared to \textbf{Rand} (\textbf{\#4}), RePro achieves clear improvements on most metrics for both Novel-split and All-splits.%, especially SGCls and PredCls.

\textbf{Ablations on VidOR}. 
We conducted the same ablation studies on VidOR-val, as shown in Table~\ref{table:VidOR-val2}. Firstly, we can observe that the compositional prompt representation shows its efficiency on both Novel-split and All-splits, \eg, 0.86\%$\rightarrow$1.72\% and 9.49\%$\rightarrow$10.06\% on R@50 of SGCls. For the motion-based prompt groups, the improvement of RePro is small due to the biased data distribution~\citep{li2021interventional}, \ie, the predicate categories strongly depend on the visual cues of subject and object tracklets, making the model predict relations simply based on object appearances without considering motion cues. More results on VidOR are left in the Appendix (Sec.~\ref{App_VidOR}).

\section{Conclusions}
% In this paper, we introduced a challenging video visual relation detection task on open-vocabulary data --- Open-VidVRD. We analyzed two key characteristics, \ie, compositional and motion-related, when applying the method of prompt tuning in this new setting. We proposed a novel method called RePro that learns compositional prompt representations while considering motion-based contexts in the video. Our evaluations on both conventional and open-vocabulary datasets show a clear superiority of RePro for tackling video visual relation detection tasks.
In this paper, we introduced the challenging Open-VidVRD task. We analyzed two key characteristics, \ie, compositional and motion-related, when applying prompt tuning in this new task. We proposed a novel method called RePro that learns compositional prompt representations while considering motion-based contexts. Our evaluations on both conventional and open-vocabulary datasets show a clear superiority of RePro for tackling video visual relation detection tasks.
$\qquad$
{\footnotesize \textbf{Acknowledgement:} This work was supported by the National Key Research \& Development Project of China (2021ZD0110700), the National Natural Science Foundation of China (U19B2043, 61976185), and the Fundamental Research Funds for the Central Universities (226-2022-00051). This work was also supported by A*STAR under its AME YIRG grant (Project No. A20E6c0101), and Singapore MOE Tier 2.}

% In this paper, we introduced the challenging Open-VidVRD task, and analysed its two key characteristics for the prompt tuning. Consequently, we proposed our RePro which consists of compositional and motion-based multi-mode prompts design. Experiment results show that our RePro achieves SOTA performance on two VidVRD benchmarks with only the base category training samples. Extensive ablations also demonstrate the effectiveness of the prompt design.$\quad$
% \textbf{Future Work}. We can design other fancier motion capturing approaches, \eg, automatically learning the motion primitives from the training set. Then for each test sample, the motion pattern can be decomposed as the weighted combination of motion primitives. Correspondingly we used the weighted combination of the prompt representations as the desired prompt representation. %We leave this as future work.

\section*{Ethics and Reproducibility statements}

\textbf{Ethics statement}.
The open-vocabulary video visual relation detection (Open-VidVRD) that we introduced in this paper is a general extension of conventional VidVRD, and there are no known extra ethical issues in terms of the Open-VidVRD task and the proposed RePro model. As for the pre-trained visual-language model (VLM) used for Open-VidVRD, the large-scale pre-training data might contain some videos and captions involved with discrimination/bias issues. When applying pre-trained VLM to Open-VidVRD, the model tends to predict relations based more on the pre-trained knowledge, and focus less on the visual cues. For example, when the pre-trained data involved with unethical videos and captions, a model might predict ``person-punch-dog" given a video of person caressing dog, which implies a person is abusing animals. To avoid the potential ethical issues, we can design algorithms to filter out those unethical training data for VLMs. For Open-VidVRD models, we can also introduce some common sense knowledge and design some rule-based methods to filter out those unreasonable relation triplets that involve ethical issues. 

\textbf{Reproducibility Statement}.
Our RePro is mainly implemented based on the realsed code of ALPro~\citep{li2022align}, VinVL~\citep{zhang2021vinvl}, and VidVRD-II~\citep{shang2021video}. We first modified the code of VinVL to fit the video data and to extract object tracklets in each video. Then We modified the code of VidVRD-II to be compatible with the visual and text encoder of ALPro, and to fit the Open-VidVRD setting. We provide the the detailed base/novel split information of object and predicate categories in the Appendix (Sec.~\ref{App_split}) to ensure all experiments can be reproduced. When training the RePro model and its variants, we manually set the random seed and fixed the seed for all experiments to ensure they can be reproduced. We also provide the code of our RePro model and the training/evaluate scripts in the supplementary materials.

\bibliography{OpenVocVidVRDbib}
\bibliographystyle{iclr2023_conference}

\appendix
\section{Appendix}

This Appendix has the following contents:
\begin{itemize}[leftmargin=15pt]
    \item More details about the relative position feature are in Sec.~\ref{App_pos}.
    \item More details about the motion pattern are in Sec.~\ref{App_motion}.
    \item \Rd{More details about the hyperparameters are given in Sec.~\ref{app_hyperparameters}.}
    \item \Rd{Analysis of the performance improvement in different predicate groups are in Sec.~\ref{App_pred_group}.}
    \item \Rd{Potential improvements of the motion pattern design are introduced in Sec.~\ref{app_futurework}.}
    \item \Rd{The detailed experiment settings of the compared SOTA methods are introduced in Sec.~\ref{app_sota}.}
    \item More experiment results on VidOR are provided at Sec.~\ref{App_VidOR}.
    \item The detailed base/novel split information of object and predicate categories are in Sec.~\ref{App_split}.
\end{itemize}

\subsection{Relative Position Feature for Tracklet Pairs}\label{App_pos}
We compute the relative position between bounding boxes of subject-object tracklet pair ⟨$T_i,T_j$⟩ by following \cite{shang2021video}. Specifically, we compute the relative position feature between subject-object for the beginning and ending bounding boxes of their temporal intersection. For the beginning bounding boxes of the subject and object, the position feature is calculated as:
\begin{align}
    \bm{f}_{i,j}^B = \left[
        \frac{x_i-x_j}{x_j},\frac{y_i-y_j}{y_j},\log\frac{w_i}{w_j},\log\frac{h_i}{h_j},\log\frac{w_ih_i}{w_jh_j},\frac{t_i-t_j}{L_\text{seg}}.
    \right],
\end{align}
where $(x_i,y_i)$ is the central coordinates of $T_i$'s beginning bounding box, $(w_i,h_i)$ is its width and height, and $t_i$ is the frame ID of this beginning bounding box. $(x_j,y_j,w_j,h_j,t_j)$ is defined similarly for $T_j$. $L_\text{seg}$ is the number of frames in each video segment, and following \cite{shang2021video}, we set $L_\text{seg}=30$. The relative position feature between the ending bounding boxes of ⟨$T_i,T_j$⟩ is defined similarly as $\bm{f}_{i,j}^B$, and is denoted as $\bm{f}_{i,j}^E$. The final relative position of ⟨$T_i,T_j$⟩ is concatenated as $\bm{f}^s_{i,j} = [\bm{f}_{i,j}^B,\bm{f}_{i,j}^E]$, and $\bm{f}^s_{i,j} \in \mathbb{R}^{12}$.

\subsection{Details About the Motion Patterns}\label{App_motion}

\begin{figure}[h]
    \centering
    % \vspace{-2ex}
    \includegraphics[width=\linewidth]{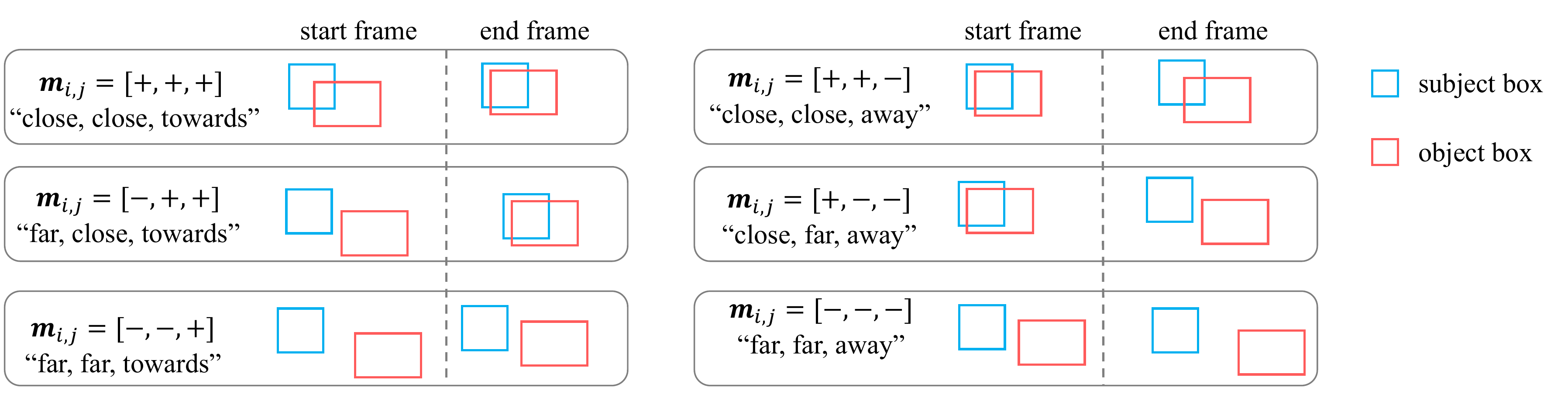}
    \vspace{-2ex}
    \caption{The schematic of the 6 motion patterns defined in Eq.~(\ref{Eq_motion}).}
\label{fig_motion}
% \vspace{-3ex}
\end{figure}

We provide a schematic of the motion patterns defined in Eq.~(\ref{Eq_motion}), as shown in Figure~\ref{fig_motion}. According to the definition in Eq.~(\ref{Eq_motion}), $\bm{m}_{i,j}$ has only 6 possible values, \ie, $[+,-,+]$ and $[-,+,-]$ are impossible.

\Rd{
\subsection{Hyperparameters}\label{app_hyperparameters}

% \textbf{Relation Detection Details}. Following the popular segment-based methods~\citep{qian2019video,shang2017video,shang2021video}, we first detected visual relations in short video segments, and then adopted greedy relation association algorithm~\citep{shang2017video} to merge the same relation triplets. 
We set $\phi_o, \phi_p$ and $\phi_\text{pos}$ all as two-layer MLPs with hidden dimension 768. The $\lambda$ for weighting the distillation loss (\ie, Eq.~(\ref{eq_l1})) was set as 5.0. The prompt length $L$ was set as 10. The softmax temperature $\tau$ was set as learnable. The GIoU threshold $\gamma$ was chosen based on the statistics of the training set, by making the tracklet pairs evenly distributed w.r.t different motion patterns. In our implementation, $\gamma$ was set as -0.3 for VidVRD and -0.25 for VidOR. We trained our RePro using Adam~\citep{kingma2014adam} with a learning rate 1e-4, and stopped the training when SGDet mAP drops.
}

\Rd{
\subsection{Analysis of the performance improvement in different predicate groups}\label{App_pred_group}

We evaluated the Recall@100 in the PredCls setting of some predicate groups (grouped by the prefix of predicate words) at the novel-split of the VidVRD dataset. We compared our RePro with the mean ensemble (Ens) and random select (Rand) variants of RePro (refer to Sec.~\ref{sec_ablation}). The results show that the improvements of motion-related predicates are much larger than other context-related predicates. For example, we have 9.68\% absolute improvements on ``run" (e.g., ``run past", ``run next to"). While for those predicates that can be roughly inferred by the context (e.g., ``fly", ``swim"), our approach has limited contributions. This indicates that the performance improvements of our RePro are largely attributed to motion cues.

}

\begin{table}[h]
        % \vspace{-1ex}
        \caption{\Ra{Recall@100 (\%) of PredCls on the test set of VidVRD w.r.t different predicate groups.}}
        \label{table:VidOR-val-appendix}
        % \vspace{-1ex}
        \centering
        % \addtolength{\tabcolsep}{-1pt}
        \begin{tabular}{ccccccccc}
        \hline
        \Ra{Methods} & \Ra{move}           & \Ra{sit}            & \Ra{run}            & \Ra{walk}           & \Ra{stop}           & \Ra{stand}          & \Ra{fly}            & \Ra{swim}          \\ \hline
        \Ra{Ens}     & \Ra{34.48}          & \Ra{50.92}          & \Ra{12.90}          & \Ra{18.30}          & \Ra{37.03}          & \Ra{35.51}          & \Ra{37.50}          & \Ra{\textbf{15.38}} \\
        \Ra{Rand}    & \Ra{37.93}          & \Ra{51.85}          & \Ra{16.12}          & \Ra{18.30}          & \Ra{\textbf{44.44}} & \Ra{36.44}          & \Ra{\textbf{50.00}} & \Ra{\textbf{15.38}} \\
        \Ra{RePro}   & \Ra{\textbf{44.82}} & \Ra{\textbf{55.55}} & \Ra{\textbf{25.80}} & \Ra{\textbf{18.95}} & \Ra{40.47}          & \Ra{\textbf{41.12}} & \Ra{\textbf{50.00}} & \Ra{12.82}          \\ \hline
        \end{tabular}
\end{table}

\Rd{
\subsection{Potential Improvements of the Motion Pattern Design}\label{app_futurework}

The current proposed GIoU-based motion pattern design can be further improved. Based on our proposed motion-pattern based prompt group learning framework, we can design other fancier motion capturing approaches, \eg, automatically learning the motion primitives from the training set. Then for each test sample, the motion pattern can be decomposed as the weighted combination of motion primitives. Consequently we can use the weighted combination of the prompt representations as the desired prompt representation. We leave this as future work.
% we decomposed its motion pattern as the weighted combination of motion primitives. Correspondingly we used the weighted combination of the prompt representations as the desired prompt representation. We leave this as future work.  
}
\Ra{
\subsection{Detailed Experimental Settings of The Compared SOTA Methods}\label{app_sota}
Since VidVRD is a very challenging task, The object tracklets and features used in SOTA methods in Table~\ref{table:VidVRD-conventional-sota} are not uniform. The object tracking algorithms for tracklet generation include Seq-NMS~\citep{han2016seq} and deepSORT~\citep{wojke2017simple}. The features include RoI Aligned features, I3D features~\citep{carreira2017quo}, and improved dense trajectory (iDT) features~\citep{shang2017video}  Here we enumerate their details as follows:

\begin{itemize}[leftmargin=15pt]
    \item \cite{su2020video} uses Seq-NMS for tracklets generation, and improved dense trajectory (iDT) feature and relative motion feature of tracklet pairs for relation classification.
    \item \cite{liu2020beyond} uses deepSORT for tracklets generation, and uses RoI feature, I3D feature, and relative motion feature of tracklet pairs for relation classification.
    \item \cite{li2021interventional} uses Seq-NMS for tracklets generation, and uses RoI feature and relative motion feature of tracklet pairs for relation classification.
    \item \cite{gao2022classification} uses deepSORT for tracklets generation, and uses RoI feature, I3D feature for relation classification.  
    \item Our RePro uses Seq-NMS for tracklets generation, and uses RoI feature and relative motion feature of tracklet pairs for relation classification.
\end{itemize}

}
% \textcolor{blue}{

\subsection{More Experiment Results on VidOR}\label{App_VidOR}

\begin{table}[h]
        % \vspace{-1ex}
        \caption{Performance (\%) on the validation set of VidOR.}
        \label{table:VidOR-val-appendix}
        % \vspace{-1ex}
        \centering
        \addtolength{\tabcolsep}{-2.5pt}
        \begin{tabular}{lcccccccc}
            \hline
            \multirow{3}{*}{Methods} & \multicolumn{4}{c}{Novel-split}                                                                         & \multicolumn{4}{c}{All-splits}                                                                           \\
                                     & \multicolumn{2}{c}{SGCls}                            & \multicolumn{2}{c}{PredCls}                          & \multicolumn{2}{c}{SGCls}                            & \multicolumn{2}{c}{PredCls}                          \\
                                     & \multicolumn{1}{c}{R@50} & \multicolumn{1}{c}{R@100} & \multicolumn{1}{c}{R@50} & \multicolumn{1}{c}{R@100} & \multicolumn{1}{c}{R@50} & \multicolumn{1}{c}{R@100} & \multicolumn{1}{c}{R@50} & \multicolumn{1}{c}{R@100} \\ \hline
            ALPro                    & 3.17                     & 3.74                      & 8.35                     & 9.79                      & 0.95                     & 1.32                      & 2.61                     & 3.66                      \\ \hline
            VidVRD-II                & 1.44                     & 2.01                      & 4.32                     & 4.89                      & 9.40                     & 12.78                     & 24.81                    & 34.11                     \\
            RePro$^\dagger$                  & 1.72                     & 2.30                      & 6.62                     & 8.06                      & 8.88                     & 11.52                     & 23.84                    & 31.57                     \\
            RePro                   & \textbf{2.01}            & \textbf{2.30}                     & \textbf{7.20}            & \textbf{8.35}             & \textbf{10.03}                    & \textbf{12.91}                     & \textbf{27.11}           & \textbf{35.76}                     \\ \hline
            \end{tabular}
\end{table}

We provide more experiment results for our RePro on the validation set of VidOR, as shown in Table~\ref{table:VidOR-val-appendix}. We first compare our RePro with using ALPro directly perform relation classification as Eq.~(\ref{Eq_argmax}). We find that ALPro performs slightly better than RePro on novel-split, because ALPro doesn't have the trend of fitting base categories. However, ALPro performs much worse than RePro in All-splits due to not trained on base categories. Furthermore, we also compare RePro with the VidVRD-II~\citep{shang2021video} baseline and the variant RePro$^\dagger$. We can observe that RePro outperfoms both VidVRD-II and RePro$^\dagger$ by a large margin on both novel-split and all-split.

\subsection{Detailed base/novel categories of object and predicate for VidVRD and VidOR}\label{App_split}

We list the base/novel categories of object and predicate for training and evaluating our RePro and other baselines in all experiments. We also provide more statistics information in the supplementary materials

\textbf{VidVRD Object}

25 base object categories:
\begin{itemize}
    \item[]  ``airplane", ``bicycle", ``bird", ``bus", ``car", ``dog", ``domestic\_cat", ``elephant", ``hamster", ``lion", ``monkey", ``rabbit", ``sheep", ``snake", ``squirrel", ``tiger", ``train", ``turtle", ``whale", ``zebra", ``ball", ``frisbee", ``sofa", ``skateboard", ``person"
\end{itemize}

10 novel object categories
\begin{itemize}
    \item[]   ``horse", ``watercraft", ``giant\_panda", ``fox", ``red\_panda", ``cattle", ``motorcycle", ``bear", ``antelope", ``lizard"
\end{itemize}

\textbf{VidVRD Predicate}

71 base predicate categories:
\begin{itemize}
    \item[]  
    ``behind", ``chase", ``creep\_behind", ``creep\_beneath", ``creep\_front", ``creep\_left", ``creep\_right", ``fall\_off", ``faster", ``fly\_above", ``fly\_next\_to", ``fly\_past", ``fly\_toward", ``fly\_with", ``follow", ``front", ``jump\_beneath", ``jump\_front", ``jump\_left", ``jump\_next\_to", ``jump\_right", ``jump\_toward", ``larger", ``left", ``lie\_behind", ``lie\_front", ``lie\_left", ``lie\_next\_to", ``lie\_right", ``move\_behind", ``move\_beneath", ``move\_front", ``move\_left", ``move\_right", ``move\_with", ``next\_to", ``play", ``ride", ``right", ``run\_behind", ``run\_front", ``run\_left", ``run\_past", ``run\_right", ``run\_with", ``sit\_above", ``sit\_front", ``sit\_left", ``sit\_right", ``stand\_behind", ``stand\_front", ``stand\_left", ``stand\_next\_to", ``stand\_right", ``stop\_behind", ``stop\_front", ``stop\_left", ``stop\_right", ``swim\_front", ``swim\_left", ``swim\_right", ``swim\_with", ``taller", ``touch", ``walk\_behind", ``walk\_front", ``walk\_left", ``walk\_next\_to", ``walk\_right", ``walk\_with", ``watch"
\end{itemize}

61 novel predicate categories:
\begin{itemize}
    \item[] ``above", ``away", ``beneath", ``bite", ``creep\_above", ``creep\_away", ``creep\_next\_to", ``creep\_past", ``creep\_toward", ``drive", ``feed", ``fight", ``fly\_away", ``fly\_behind", ``fly\_front", ``fly\_left", ``fly\_right", ``hold", ``jump\_above", ``jump\_away", ``jump\_behind", ``jump\_past", ``jump\_with", ``kick", ``lie\_above", ``lie\_beneath", ``lie\_inside", ``lie\_with", ``move\_above", ``move\_away", ``move\_next\_to", ``move\_past", ``move\_toward", ``past", ``pull", ``run\_above", ``run\_away", ``run\_beneath", ``run\_next\_to", ``run\_toward", ``sit\_behind", ``sit\_beneath", ``sit\_inside", ``sit\_next\_to", ``stand\_above", ``stand\_beneath", ``stand\_inside", ``stand\_with", ``stop\_above", ``stop\_beneath", ``stop\_next\_to", ``stop\_with", ``swim\_behind", ``swim\_beneath", ``swim\_next\_to", ``toward", ``walk\_above", ``walk\_away", ``walk\_beneath", ``walk\_past", ``walk\_toward"
\end{itemize}

\textbf{VidOR Object}

50 base object categories:
\begin{itemize}
    \item[] ``adult", ``child", ``toy", ``dog", ``baby", ``car", ``chair", ``table", ``sofa", ``ball/sports\_ball", ``screen/monitor", ``cup", ``bicycle", ``guitar", ``bottle", ``backpack", ``handbag", ``baby\_seat", ``camera", ``cat", ``cellphone", ``bird", ``sheep/goat", ``laptop", ``ski", ``stool", ``watercraft", ``duck", ``bus/truck", ``bench", ``fruits", ``baby\_walker", ``horse", ``bat", ``dish", ``electric\_fan", ``kangaroo", ``motorcycle", ``lion", ``hamster/rat", ``refrigerator", ``elephant", ``faucet", ``cake", ``penguin", ``sink", ``piano", ``microwave", ``cattle/cow", ``aircraft"
\end{itemize}

30 novel object categories:
\begin{itemize}
    \item[]  ``antelope", ``vegetables", ``panda", ``rabbit", ``fish", ``train", ``snowboard", ``suitcase", ``squirrel", ``leopard", ``chicken", ``skateboard", ``traffic\_light", ``surfboard", ``camel", ``racket", ``bread", ``bear", ``oven", ``scooter", ``frisbee", ``stop\_sign", ``turtle", ``stingray", ``pig", ``crab", ``crocodile", ``toilet", ``tiger", ``snake"
\end{itemize}

\textbf{VidOR Predicate}

30 base predicate categories:
\begin{itemize}
    \item[] ``next\_to", ``in\_front\_of", ``watch", ``behind", ``away", ``towards", ``beneath", ``above", ``hold", ``lean\_on", ``speak\_to", ``ride", ``hug", ``touch", ``carry", ``hold\_hand\_of", ``bite", ``push", ``pull", ``play(instrument)", ``grab", ``release", ``pat", ``inside", ``lift", ``caress", ``point\_to", ``press", ``hit", ``use"
\end{itemize}

20 novel predicate categories:
\begin{itemize}
    \item[] ``kick", ``chase", ``wave", ``smell", ``throw", ``feed", ``kiss", ``wave\_hand\_to", ``shout\_at", ``drive", ``clean", ``lick", ``squeeze", ``shake\_hand\_with", ``get\_off", ``knock", ``cut", ``open", ``get\_on", ``close"
\end{itemize}

\end{document}